\documentclass{article}

\usepackage{arxiv}

\usepackage[utf8]{inputenc} 
\usepackage[T1]{fontenc}    
\usepackage{hyperref}       
\usepackage{url}            
\usepackage{booktabs}       
\usepackage{amsfonts}       
\usepackage{nicefrac}       
\usepackage{microtype}      
\usepackage{lipsum}
\usepackage{graphicx}
\graphicspath{ {./images/} }

\title{
Enhancing the Interpretability of Rule-based Explanations through Information Retrieval
}

\author{
Alessandro Umbrico, Luca Coraci, Francesca Fracasso, Silvia Gola, Gabriella Cortellessa\\
CNR - Institute of Cognitive Sciences and Technologies
\AND
Guido Bologna\\
University of Applied Sciences and Technologies
}

\begin{document}
\maketitle
\begin{abstract}
The lack of transparency of data-driven Artificial Intelligence techniques limits their interpretability and acceptance into healthcare decision-making processes. We propose an attribution-based approach to improve the interpretability of Explainable AI-based predictions in the specific context of arm lymphedema's risk assessment after lymph nodal radiotherapy in breast cancer. The proposed method performs a statistical analysis of the attributes in the rule-based prediction model using standard metrics from Information Retrieval techniques. This analysis computes the relevance of each attribute to the prediction and provides users with interpretable information about the impact of risk factors. The results of a user study that compared the output generated by the proposed approach with the raw output of the Explainable AI model suggested higher levels of interpretability and usefulness in the context of predicting lymphedema risk.
\end{abstract}


\section{Introduction}
\label{sec:intro}

Artificial Intelligence (AI) technologies have shown impressive computational capabilities that can positively impact decision-making processes. Although effective, the black-box nature of data-driven AI systems hampers their diffusion and acceptance in the real world, especially in high-stakes decision-making domains such as healthcare.
Adopting AI in clinical practice can offer potential advantages \cite{vellido2020}, such as augmenting clinicians' work in the diagnostic phase, or providing valuable personalized treatment recommendations. Nevertheless, several issues still exist that prevent the regular adoption of such systems in the usual practice.

Lack of transparency is one of the main barriers \cite{ghassemi2021}.  Since clinicians need to be confident in their delicate role, AI systems should be trustworthy, and Explainable AI (XAI) can support the development of such systems \cite{markus2021}. 
Explainability is closely related to understanding, which involves forming a mental model of what is being observed. Explaining or interpreting can be described as the process of presenting the causes behind observed phenomena in a clear and comprehensible way, using linguistic descriptions to outline their logical and causal relationships \cite{Holzinger2019,Langer2021}.
Considering the health domain and the clinical literature, it has been argued that clinicians may feel uncomfortable with recommendations coming from black-box predictive models. AI outcomes should be explainable so that clinicians can understand the underlying (hidden) {\em rationale} \cite{Wang2019}.  Clinicians view explainability as a means of justifying health-related decisional processes in the context of model prediction \cite{tonekaboni2019}.

Within XAI, explainability aims to create insight into how and why AI models produce predictions while maintaining high predictive performance levels. The added value of explainable AI methods remains to be proven in practice \cite{markus2021}, and no final agreement has yet been reached on the definition of ``explainability'' or ``interpretability''. The work \cite{xai2023} proposes a distinction between the two terms by considering different perspectives of explanation needs. Explainability provides insights to a target audience to fulfill a need. Interpretability is the degree to which the provided insights can make sense for the target audience’s domain knowledge \cite{xai2023}.
A similar distinction is discussed in \cite{MONTAVON2018}, where an explanation is understood as a collection of features from the interpretable domain that contributed to a specific example in producing a decision. In contrast, an interpretation is defined as the mapping of an abstract concept into a domain that a human expert can perceive and understand.

Under this premise, it has been suggested that current explainability methods still function as black-boxes from a user’s perspective, as they remain difficult for non-experts. They are indeed better regarded as tools for developers and auditors to analyze their models. Despite the significant effort made in XAI, the proposed solutions should be justified by their reliability and experimentally validated performance \cite{ghassemi2021}. 
%
%
Explainable AI models do not automatically imply the {\em understandability} of their outcomes to all users (stakeholders). It is necessary to consider the explanation needs, objectives, interests, and expertise of stakeholders and communicate the results of AI models accordingly. To this aim, it is important to rely on conceptual frameworks that allow designers/researchers to think about the quality of XAI from multiple perspectives  (e.g., understandability, interpretability, usability, usefulness) \cite{combi2022}.

The current work concerns the enhancement of an XAI model designed to predict the risk of arm lymphedema in breast cancer treatments \cite{bologna2024}. Arm lymphedema can represent a side effect with a significant impact on the quality of life of patients. The ability to provide a personalized prediction of the risk of lymphedema enables more personalized treatment planning and the provision of lifestyle advice to minimize the risk and severity of this side effect.
%
%
In this complex domain, a key challenge lies in the inherent complexity of AI models, which often makes their outputs difficult to understand for non-expert users, including clinicians and patients. The results of AI models such as \cite{bologna2024} that are considered ``explainable'' by AI experts, even though they are effective, are far from being fully understood by users without an advanced AI background.
%

We propose an approach that infers causal correlations between the predictions' outcomes of an explainable rule-based AI model \cite{bologna2024} and its input attributes without affecting the underlying learning mechanism (i.e., we consider the AI model as a black-box). The approach aggregates input variables into semantically coherent {\em contributing factors} and proposes {\em rankings} characterizing the {\em contribution} of factors within each prediction. The contribution of each factor is computed by taking inspiration from Information Retrieval (IR) principles, specifically the \texttt{tf-idf} metric. 
We show that this analysis enhances the explainability of a rule-based (explainable) AI model and provides users without an AI background with useful insights into the interpretation of the predictions. In particular, our approach allows users to characterize the different contributions of factors to risk prediction (i.e., AI outcome) and, consequently, effectively assess the quality of the AI model. We describe an experimental analysis showing the efficacy of the designed approach in contextualizing the contributions of risk factors to different profiles and the increased interpretability of the explainable model.

\section{Explanation Needs in Risk Prediction and Communication}
Research initiatives are investigating explanation and communication techniques that are suitable to effectively interpret lymphedema risk in breast cancer patients who should be treated with loco-regional radiotherapy \cite{preact}. Such initiatives aim to address several of the open challenges concerning explainability for ``clinicians'' \cite{ghassemi2021,amann2020}. 
%
Considering the complexity of the domain and the challenging objective of addressing the explanation needs of clinicians and patients, we rely on an AI approach supporting explainability by ``construction'' through rule-based predictions \cite{bologna2024}.

Predictions are expressed as logical conjunctions among numeric conditions on the model's attributes. Specifically, the prediction model is an ensemble of Multi-Layer Perceptrons (MLPs). Rules from the (MLP) ensembles are generated by leveraging a special property of a particular MLP model, whose discriminant boundaries are parallel to the axis of the input variables. In addition, these discriminant hyperplanes are precisely located.
Knowing the exact location of the axis-parallel hyperplanes, an algorithm has been presented that, given a sample, determines a propositional rule. 
The rule antecedents correspond to the hyperplanes. At each step, the rule extraction algorithm computes the best antecedent according to a fidelity-based criterion, which corresponds to how a rule mimics the behavior of a model. For example, with $s$ samples that activate a rule and $s'$ samples for which the classifications of that rule match the classifications of the model, the fidelity is $s'/s$. A second rule extraction algorithm has been developed that aims to generate a set of rules for a training set of size $S$.
It corresponds to a covering technique that calls the previous algorithm $S$ times. It thus generates $S$ rules and heuristically selects a subset that covers all $S$ samples \cite{bologna2024}.

The logical representation explicitly correlates input variables' values with the conditions that determine the outcomes. Such rules are generally considered {\em explainable} thanks to their logical representation \cite{markus2021}. However, domain experts like doctors or patients might not fully understand the rules. 
To improve the explainability of this model, it is necessary to further elaborate rules and provide users with alternative (and more abstract) views of the predictions. Namely, it is necessary to support different {\em interpretations} of the predictions to better communicate predicted risks to target users (i.e., doctors, patients, model developers). 
On the one hand, certain details of the rules are useful to AI experts to validate and check the model. On the other hand, some details are not necessary for doctors or patients to understand the {\em meaning} of the predictions.

\section{Beyond Rule-based Explanations}
The proposed approach aggregates attributes of the AI module into coherent clusters that correspond to semantically coherent factors. Given the mapping between the attributes of the model and the factors, the designed XAI service elaborates the rules and computes the contribution of each factor on the prediction.
The relevance of a factor is computed through IR metrics \texttt{tf} ({\em term frequency}), \texttt{idf} ({\em inverse document frequency}), and \texttt{tf-idf}. The computation determines: (i) the (sub)set of factors that actively affect the prediction of a patient, and (ii) to what extent each active factor determines the outcome of the prediction (i.e., its {\em relevance}).

The use of IR metrics is largely inspired by the fact that the considered rule-based model \cite{bologna2024} makes predictions by activating an arbitrary number of rules for a given input (i.e., patient data). Each rule of the model is characterized by a conjunction of conditions on a subset of the input attributes (and factors). Therefore, an attribute could not occur at all in the rules activated for a given input (i.e., the attribute does not determine the predicted outcome) or can occur multiple times (in several of the activated rules).
Several rules can be activated for a given input (i.e., for each prediction of a patient). The underlying assumption is that the more a set of attributes is present in the precondition of the activated rules, the greater the relevance (or contribution) of the associated factor on the considered prediction.
Therefore, the inspiring ``metaphor'' is that attributes, likewise tokens in documents, could appear zero or multiple times in a rule. If we see a prediction as a collection of rules, it is possible to assess the relevance of attributes (and associated contributing factors) to a prediction, in the same way IR assesses the relevance of tokens to a collection of documents.

Recalling IR concepts, the objective of \texttt{tf-idf} is to infer the amount of information conveyed by a token to classify the content of documents within a collection. The metric generally assigns higher values to tokens that are not widely used over the collection but are specific to a subset of documents (i.e., tokens that are specific to only a limited number of documents within the collection). The assumption is that tokens that are not common but frequent only in some documents are more specific and thus better capture the content of those documents. 
To compute the relevance of factors in a collection of rules, we interpret each rule as a document and each factor as a token that can be present in a document. Factors are counted according to the occurrence of the associated attributes in the rules' conditions. Unlike the typical IR applications, we are interested in the factors that obtained a low \texttt{tf-idf} score, meaning that such factors appear in many activated rules and therefore are those that influence the prediction the most.


%
%
%
\subsection{Semantic Clustering of Attributes}
\begin{figure}[ht!]
    \centering
    \includegraphics[width=0.7\linewidth]{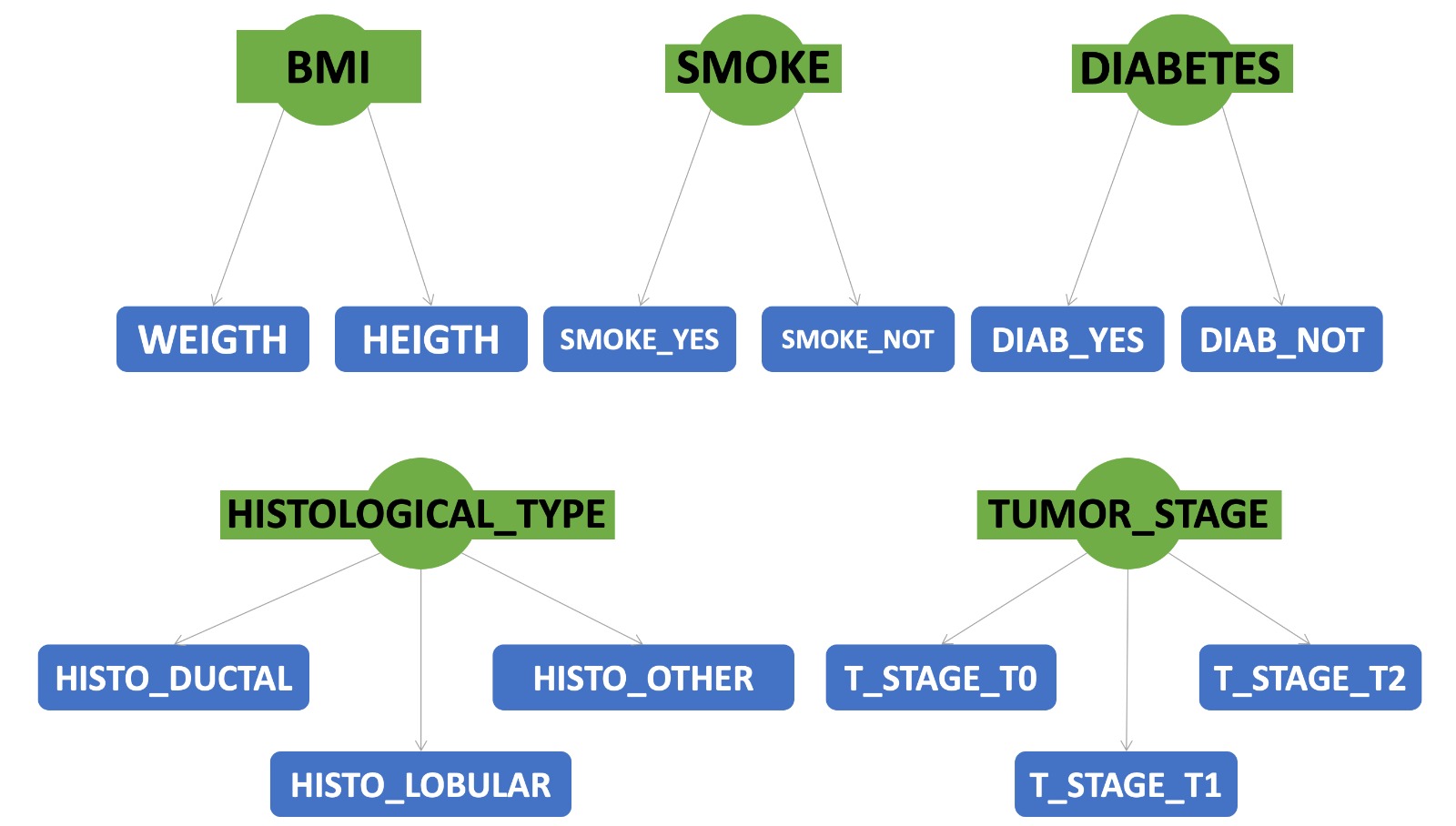}
    \caption{Excerpt of the mapping between attributes and factors.}
    \label{fig:taxonomy}
\end{figure}

It should be noted that the proposed interpretation methodology is flexible to clustering. Namely, it is possible to consider different clustering with different levels of detail on the set of variables of the prediction model. Such flexibility is well suited to personalize the explanation (and consequently the interpretation process) to the profile or expertise of the target user.
In general, clustering determines a taxonomical structure linking the attributes of the AI model to contributing factors. Figure \ref{fig:taxonomy} shows an excerpt of a taxonomical structure. The nodes in the upper part are the contributing factors, while the nodes in the lower part are the attributes of the model. The edges pointing from the top (the factors) to the bottom (the attributes) describe the aggregation resulting from the clustering.
An example of clustering is the mapping of the model attributes \texttt{SMOKER}, \texttt{FORMER SMOKER}, \texttt{CURRENT SMOKER} to the factor \texttt{SMOKER}. Or, the mapping of the model attributes \texttt{IMRT CHM}, \texttt{IMRT TOMOTHERAPY}, \texttt{IMRT X3D}, \texttt{IMRT VMAT} to the factor \texttt{RT TECHNIQUE}.

The granularity level of the aggregation determines the level of detail of the consequent computation of the relevance of the factors. Different levels of granularity (i.e., information details) are necessary to effectively communicate the risk to different stakeholders (e.g., doctors, AI experts, patients). The clustering can thus tailor the interpretation by adapting the level/amount of information conveyed to the {\em recipient of the explanation}, according to the expected expertise. For example, AI experts may prefer a {\em technical interpretation} showing a high level of detail to assess/debug the model. They might prefer a 1:1 mapping between the factors and the attributes. A doctor would instead prefer a {\em clinical interpretation} aggregating attributes in a clinically coherent way to assess the clinical quality of the prediction.
The clustering associated with the {\em clinical interpretation} is the one used within the current work to assess the interpretability of the proposed approach.

\subsection{Factor Impact Computation}
Each factor is a set of clinically coherent attributes of the model that appear in the preconditions of the rules \footnote{In case of acceptance, we will publish the documented open-source repository and the REST API endpoint exposing a XAI engine implementing the proposed logic.}. 
Multiple attributes of the same cluster/factor might appear in the preconditions of a rule. So, considering $n_{ij}$ as the number of attributes related to the factor $i$ that appear in the rule $j$, and $d_j$ as the number of attributes in the rule, the term frequency is:

\begin{equation}
    tf_{ij} = \frac{n_{ij}}{d_j}
\end{equation}

\noindent
the $tf_{ij}$ score is weighted according to the (relative) coverage of a rule. The analysis first computes the total coverage ($c_{all}$) of all the rules activated for the current patient and then computes the weight $w_j$ by considering its coverage $c_j$ as follows:

\begin{equation}
    w_j = \frac{c_j}{c_{all}}
\end{equation}

\noindent
the final term frequency metric is

\begin{equation}
    tf_{ij} = w_j \times \frac{n_{ij}}{d_j}
\end{equation}

The inverse document frequency $idf_i$ then expresses how much a factor $i$ is common within a set of activated rules. Following the standard definition of the metric, considering $s_i$ as the number of rules where the factor $i$ occurs and $s_{all}$ the number of rules in the prediction, the inverse document frequency is

\begin{equation}
\label{eq:idf}
    idf_i = \log{\frac{s_{all}}{s_i}}
\end{equation}

\noindent
the final score $tfidf_{ij}$ of a factor $i$ for a rule $j$ is then obtained as 

\begin{equation}
    tfidf_{ij} = tf_{ij} \times\ idf_i
\end{equation}

A high value of $tfidf_{ij}$ of a factor $i$ for a rule $j$ means that the factor $i$ is highly specific to the rule $j$ (high $tf_{ij}$) and not common within the prediction (low $idf_i$). Since the argument of the log function of Eq. \ref{eq:idf} is always greater than or equal to 1, the value of the resulting $idf_i$ (and $tfidf_{ij}$) is greater than or equal to 0. 
If a factor $i$ appears in many rules of a prediction, the argument inside the log function of Eq. \ref{eq:idf} approaches 1. Consequently, the associated $idf_i$ and $tfidf_{ij}$ values are close to 0.

The relevance score $tfidf_i$ of a factor $i$ within a prediction considers the average $tfidf_{ij}$ calculated over the activated rules $j$. Considering $\mathcal{J}$ as the set of rules activated for a given input, the relevance score $tfidf_i$ is thus as follows: 

\begin{equation}
    tfidf_i = \frac{\sum_{j \in\mathcal{J}} tfidf_{ij}}{|\mathcal{J}|}
\end{equation}

Unlike the standard use of such metrics in IR, we are interested in factors that are widely used in the rules of a prediction (i.e., tokens that appear in many documents of a collection). 
We ``reverse'' the typical meaning of $tfidf_i$ to assign a higher score to factors $i$ appearing in many rules, rather than to factors appearing in a few rules. To this end, we first distribute the calculated $tfidf_i$ scores within the range [0,1] through a standard logistic function and then invert the score within the interval as follows:

\begin{equation}
    tfidf^{*}_i = 1 - \frac{1}{1 + e^{-tfidf_i}}
\end{equation}

%
%
The resulting relevance score $tfidf^{*}_i$ of a factor $i$ is specified within the interval [0, 1]. Values close to 1 are assigned to factors $i$ that are common within the rules of a prediction. Values close to 0 are instead assigned to factors $i$ that are not common in the rules activated for a given input.

\section{Experimental Evaluation}
\label{sec:exps}
The experiments assess the feasibility and interpretability of the proposed approach. The first part of the assessment focuses on the technical implementation of the proposed analysis. It shows how the approach enriches both global and local explanations. The proposed approach effectively contextualizes {\em factor impact} according to different risk predictions. 
The second part of the assessment focuses on the efficacy of the generated interpretations. It analyzes feedback from users of different fields, who were categorized into AI experts (e.g., AI developers, data analysts) and non-AI experts (e.g., health professionals, clinical researchers, caregivers) that evaluate the perceived interpretability and usefulness of the information provided at different levels of interpretability \footnote{The supplementary material contains the data used for the experiments. Here, the reader can also find files containing the rules characterizing the assessed version of the model.
}.

\begin{figure}
    \centering
    \includegraphics[width=0.45\textwidth]{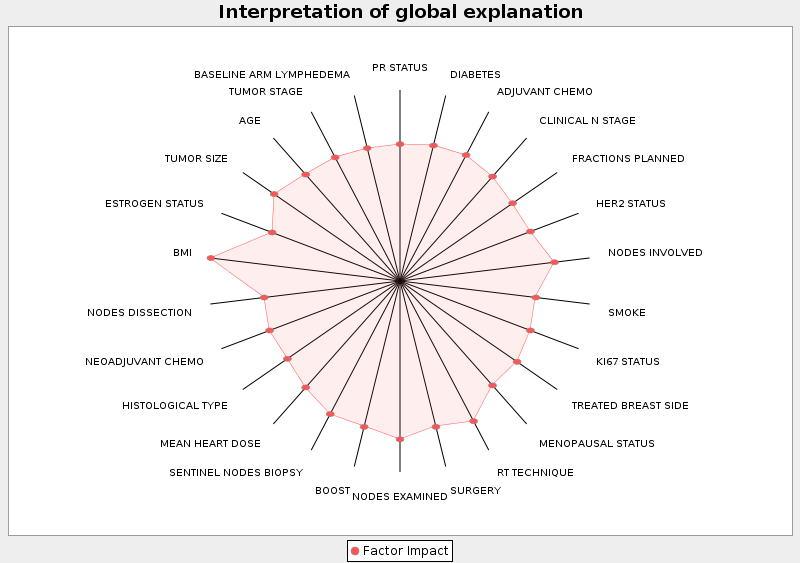}
    \caption{Interpretation of the global explanation generated on the whole set of rules (i.e., the predictive model).} 
    \label{fig:global-expl}
\end{figure}

\subsection{Global Explanation}
The first evaluation concerns the interpretability of the {\em global explanation} of the AI model. A global explanation concerns the analysis of the implicit knowledge encapsulated by all the rules of the AI module.
It is generally useful to {\em debug} the model or to support {\em knowledge discovering} \cite{markus2021}. A global explanation allows domain experts to verify the coherence of the predictive model with the literature by checking the presence of known relevant factors. It could also discover new/unexpected factors to be relevant for the prediction. This information could be useful to identify new correlations (i.e., knowledge discovery) or wrong behaviors of the model (i.e., model validation and debugging).

In our case, the model consists of 301 rules used for risk prediction. The proposed analysis computes the relevance of the risk according to the whole set of rules. Figure \ref{fig:global-expl} shows the interpretation of the global explanation, which is carried out with negligible computation time (some milliseconds).
It is important to look at the factors whose scores are higher than the average. The model implicitly recognizes such factors as particularly relevant to prediction, since they occur more frequently in the preconditions of the rules.

The factor \texttt{BMI} has the highest score and is thus the most relevant factor of the model. This is coherent with the clinical knowledge since high \texttt{BMI} is considered one of the most critical factors concerning the risk of arm lymphedema after radiotherapy . 
Other factors that receive scores higher than the average are \texttt{NODES INVOLVED} and \texttt{RT TECHNIQUES}. These factors respectively represent the number of lymph nodes that tested positive for cancer, and the type of radiotherapy treatment followed by a patient. Again, the higher relevance of these factors inferred by the model is consistent with clinical knowledge \cite{RibeiroPereira2017,park2017}.

\subsection{Local Explanations}
Local explanations concern specific patients' predictions and thus the interpretation of the subset of rules {\em activated} by the model for a given input. Patients have different clinical conditions (i.e., input values) and different contributing factors should affect the prediction. Consequently, different sets of rules are activated for each patient, leading to predictions involving different attributes (and thus factors). 
The interpretation process analyzes only the part of the model activated for a particular patient by pointing out the risk factors determining the risk prediction for a specific patient and their associated relevance.

\begin{figure*}
    \centering
    \begin{tabular}{c c c}
        \includegraphics[width=.27\textwidth]{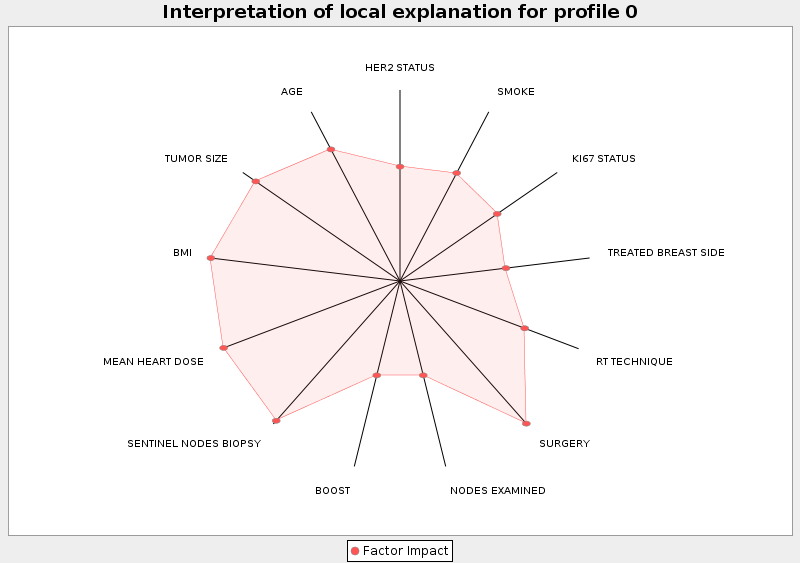} & 
        \includegraphics[width=.27\textwidth]{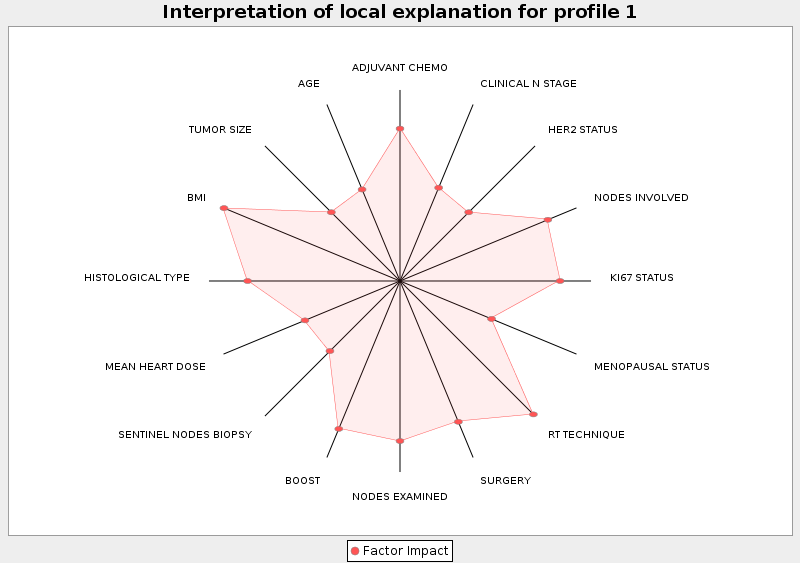} &
        \includegraphics[width=.27\textwidth]{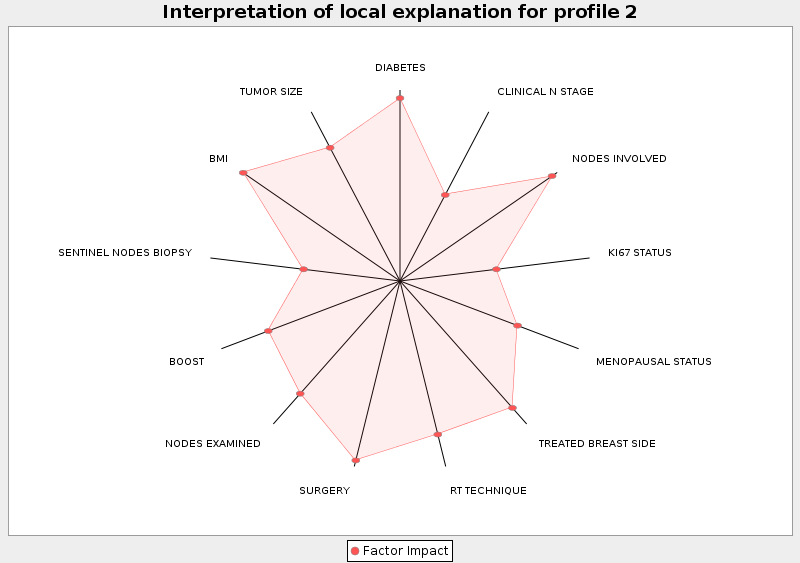}\\
        \includegraphics[width=.27\textwidth]{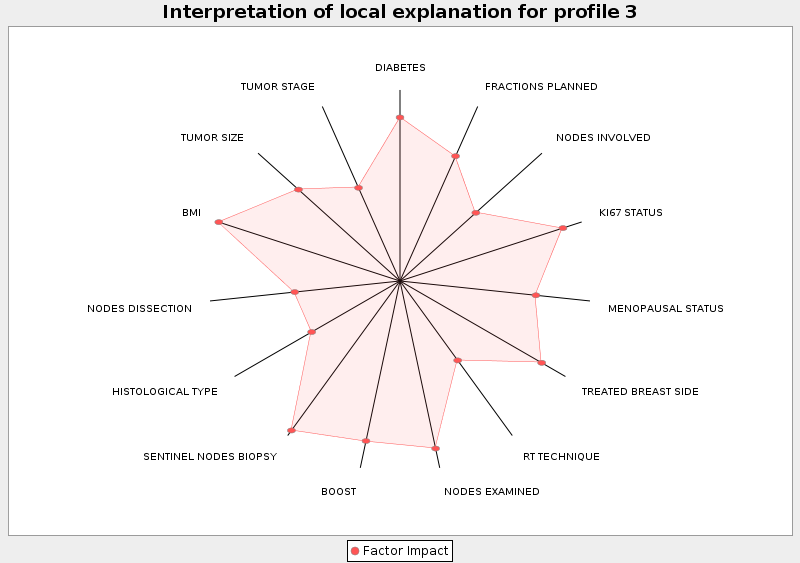} &
        \includegraphics[width=.27\textwidth]{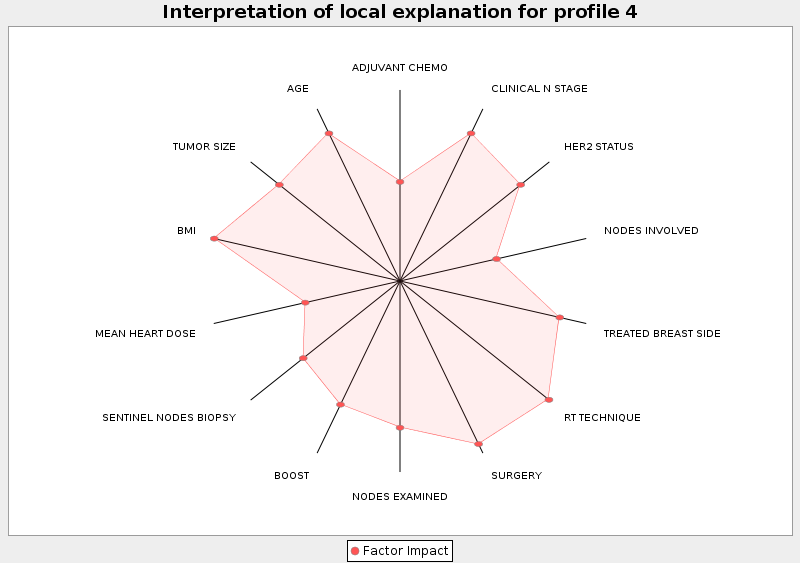} &
        \includegraphics[width=.27\textwidth]{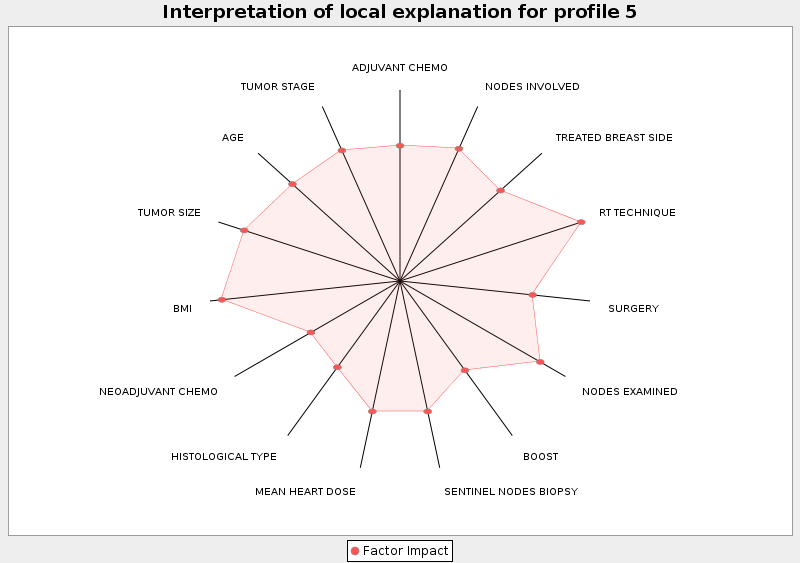}\\ 
        \includegraphics[width=.27\textwidth]{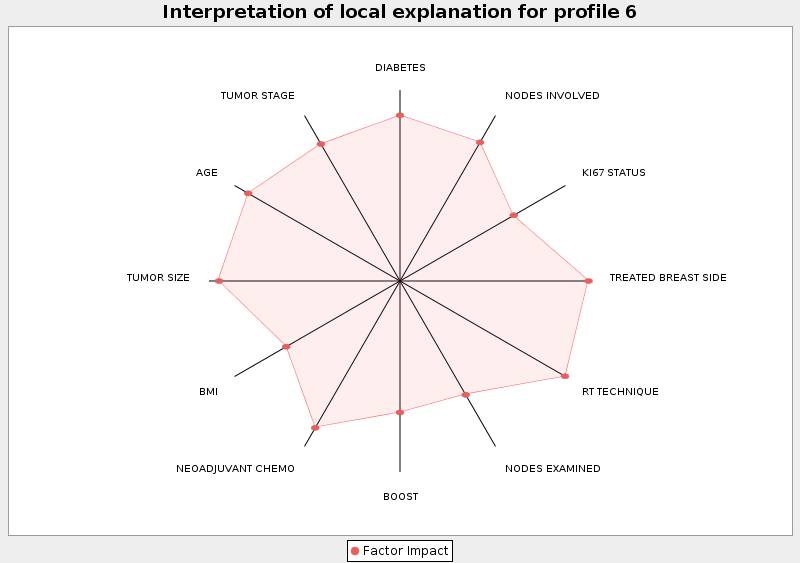} &
        \includegraphics[width=.27\textwidth]{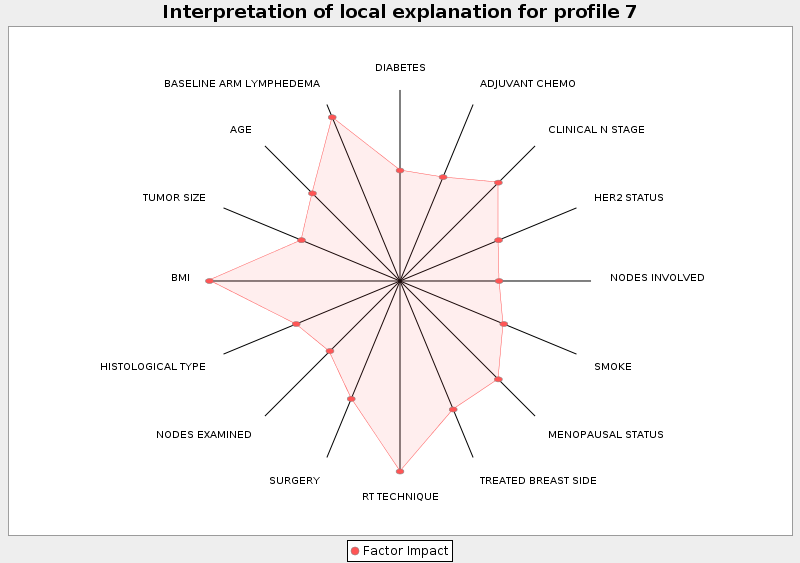} &
        \includegraphics[width=.27\textwidth]{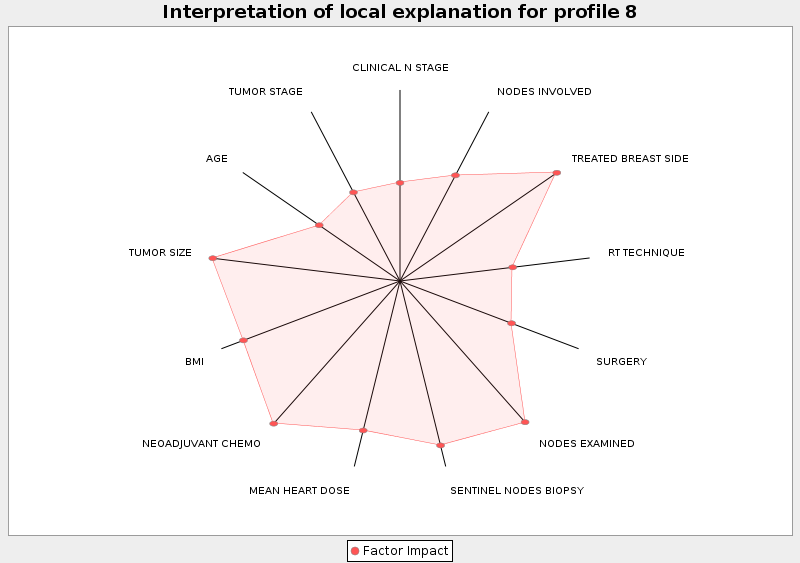}\\
        &
        \includegraphics[width=.27\textwidth]{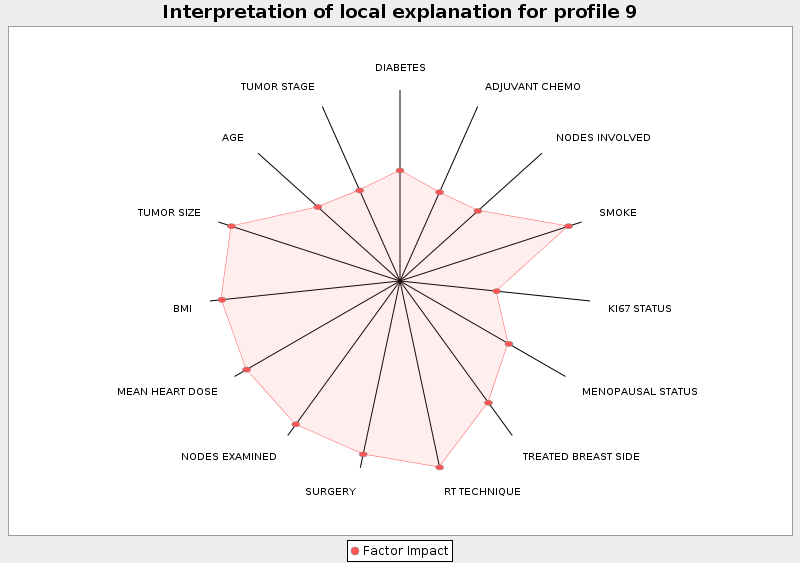} & \\
    \end{tabular}
    \caption{Interpretations of the local explanations generated for the defined patient profiles.}
    \label{fig:local-expls}
\end{figure*}

To stress the interpretation's flexibility, we consider 10 profiles with randomly assigned 7 rules each from the model. As for the global explanation, the computation time of these interpretations is negligible (milliseconds). Figure \ref{fig:local-expls} shows the interpretations obtained for these profiles. The radar charts provide insights about the {\em weights} of the involved factors in the predictions. They clearly show how the interpretation of the local explanation changes from patient to patient. The scores computed for the contributing factors indeed change according to the features of the rules activated for the predictions.

\begin{figure*}
    \centering
    \includegraphics[width=0.95\textwidth]{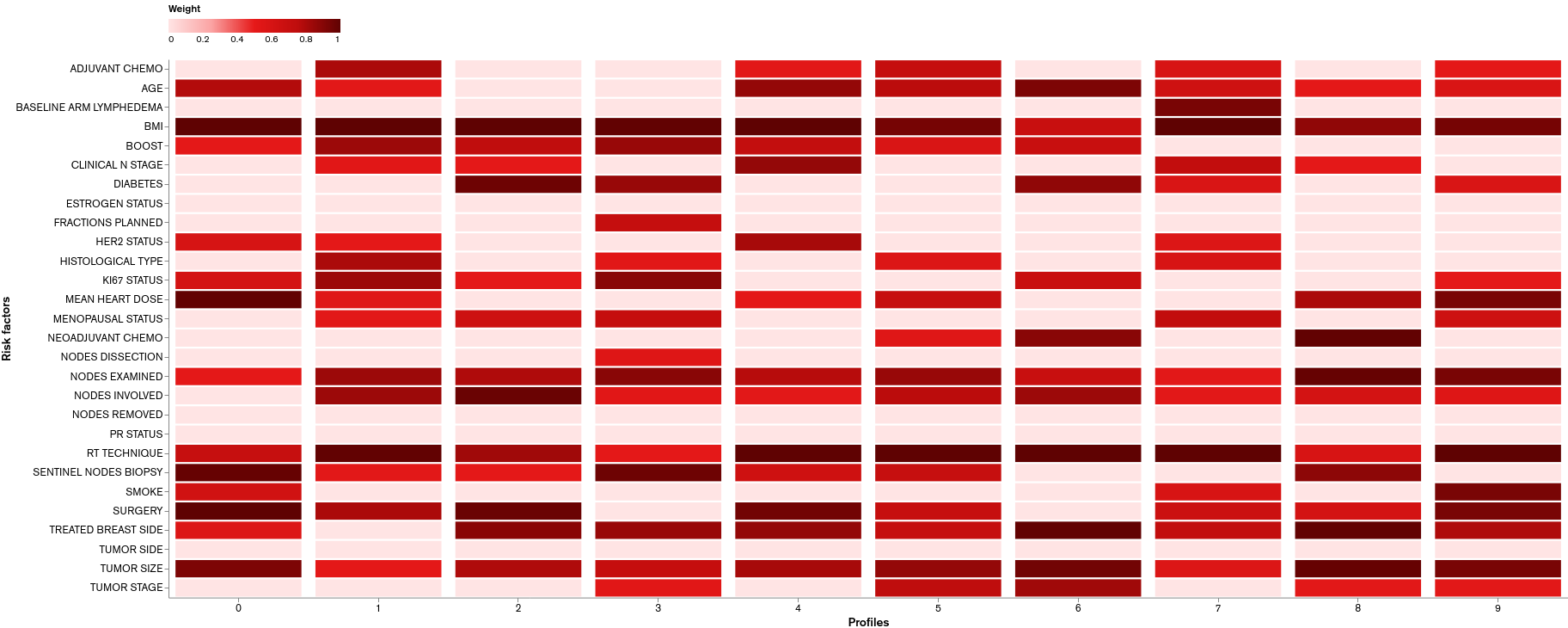}
    \caption{Heatmap showing the distribution of factor relevance over the profiles.}
    \label{fig:heatmap}
\end{figure*}

The heatmap in Figure \ref{fig:heatmap} summarizes the distribution of the scores computed for the 10 profiles. On the one hand, it clearly shows the different relevance of some factors over the considered profiles. Factors like \texttt{CLINICAL N STAGE}, \texttt{SENTINEL NODES BIOPSY}, or \texttt{ADJUVANT CHEMOTERAPY} indeed have high or low relevance depending on the profile. Some factors like \texttt{ESTROGEN STATUS}, or \texttt{PR STATUS} instead have low relevance on all profiles. Other factors like \texttt{BMI}, \texttt{RT TECHNIQUE}, or \texttt{TUMOR SIZE} have high relevance on all profiles.

\subsection{Interpretability Assessment}
To understand whether the proposed analysis and the computed relevance scores can help users interpret the output, a study was designed in which participants were asked to evaluate the XAI model's output presented through three different representations:

\begin{itemize}
    \item \textit{Raw output containing propositional rules}, as generated by the model, consisting of the estimated risk with its confidence interval, accompanied by locally activated rules representing the attributes contributing to the risk.
    \item \textit{Radar graph displaying the contributing factors}, namely the processed information using the technique presented in this paper, visualized through a radar chart where the contributing factors to the risk are represented as the spikes, and their weight is conveyed through the chart.
    \item \textit{List of factors contributing to the risk prediction}, namely the processed information using the technique from this work, presented as a list of contributing variables where their impact on the risk is represented as a percentage.
\end{itemize}

\begin{figure*}
    \centering
    \includegraphics[width=.99\textwidth]{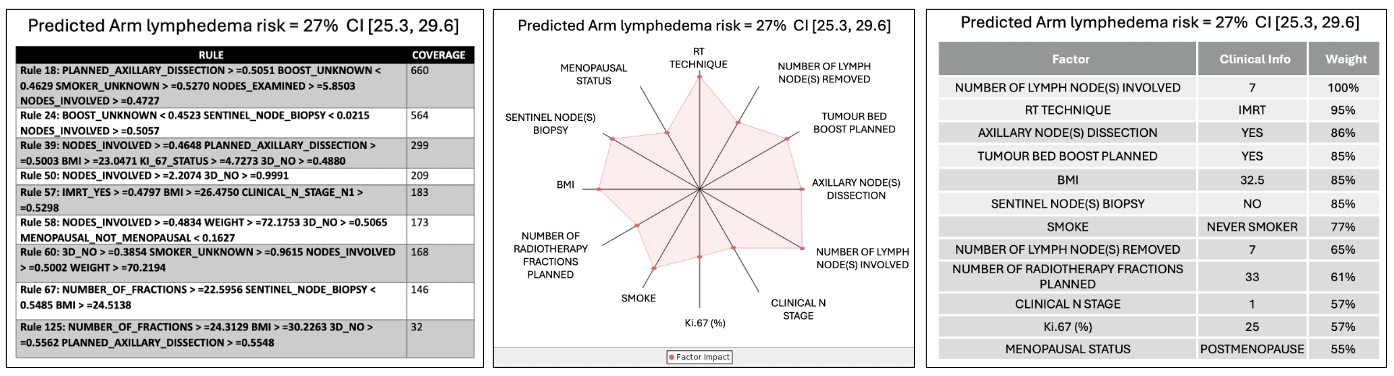}
    \caption{The three different output visualizations for a single patient profile. From left to right: Raw output containing propositional rules; Radar graph displaying the contributing factors; List of factors contributing to the risk prediction.}
    \label{fig:visualization}
\end{figure*}

\noindent
These three output visualizations (Figure \ref{fig:visualization}) were generated for four profiles corresponding to realistic patients with different risk levels. Participants in the study were asked to indicate on a 5-point Likert scale the level of \textit{interpretability} and \textit{usefulness} concerning the information conveyed for each output representation across all four profiles of patients. 

In order to estimate a valuable sample size, an a priori power analysis was conducted using G*Power \cite{Faul2009}. It indicated that, to detect an effect size of 0.6 with 80\% power and a significance level of 0.05, a sample of 28 participants was required. More than 28 participants were involved in order to face the risk of sample size reduction due to subject exclusion during the analysis (i.e. missing data, outliers).
Finally, 30 persons were recruited via convenience sampling in this evaluation phase, both AI experts and non-AI experts from different areas of expertise. No further information on the participants was collected, in line with the data minimization principle under GDPR, as stated in Article 5(1)(c) \cite{gdpr2016}. 

\begin{figure}
    \centering
    \includegraphics[width=0.9\textwidth]{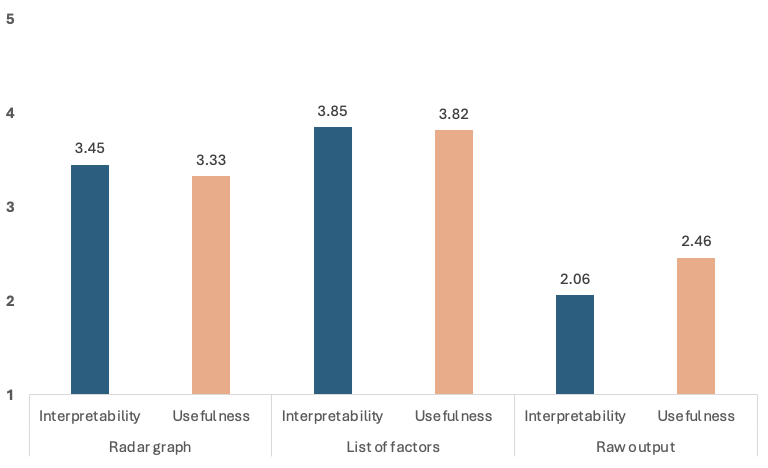}
    \caption{The mean scores obtained for both \textit{Interpretability} and \textit{Usefulness} of the three visualization options. }
    \label{fig:results}
\end{figure}

Results showed statistically significant differences in the scores provided for both \textit{Interpretability} and \textit{Usefulness} given to the three visualization options (Figure \ref{fig:results}). 
More in detail, a repeated measures ANOVA revealed a significant main effect of the different types of visualization on \textit{Interpretability}, (\textit{F\textsubscript{(2, 56)}}= 35.9; \textit{p =} .001, $\eta^2_{\mathrm{G}}$ = 0.375).  

Furthermore, post-hoc analysis showed no significant differences between the radar graph and the list of factors, while raw output interpretability scores were significantly lower when compared with both the factors' list (\textit{t\textsubscript{(28)}}= 7.74; \textit{p=} .001) and the radar graph visualization (t\textsubscript{(28)}= 7.52; \textit{p=} .001)

 The same results emerged for the \textit{Usefulness} (\textit{F(\textsubscript{(2,56)})}= 20.14; \textit{p=} .001, $\eta^2_{\mathrm{G}}$ = 0.235), and post-hoc analysis found significant differences between raw output usefulness compared to both the graphical visualization (\textit{t\textsubscript{(28)}}4.59; \textit{p=} .001) and the visualization through the list of contributing factors (\textit{t\textit{\textsubscript{(28)}}}= 5.77; \textit{p=} .001).
When comparing the obtained assessments between AI experts and non-AI experts, no significant differences were found in the judgments regarding the \textit{Interpretability} and \textit{Usefulness} of the three modalities.

Overall, the results of our user study indicate that the interpretability of the AI model’s output significantly improves when the information is further elaborated with the goal of simplifying its presentation. Notably, the enhanced visualizations were perceived as easier to understand by both AI experts and non-AI experts alike. In addition to increased clarity, participants also rated these visualizations as more useful, particularly in light of their intended practical application. This suggests that thoughtful simplification does not undermine content accuracy, but rather supports more effective communication across different user profiles.

\section{Discussion of Results and State of the Art}

The approach investigates attribution-based analysis to improve the interpretability of a rule-based prediction model.  The motivations of our approach come out from the observation that, although explainable, it is not enough for doctors and patients who are far from being AI experts. Due to privacy constraints that limited access to raw input data and the possibility of retraining the AI model, which lead us to the need of using the AI model as black-box. We addressed the interpretability model of the explainable model output by looking at the attribution-based methodology. Attribution-based explainability is a well-established methodology, generally applied to black-box models \cite{markus2021}.

An approach typically used in the literature is the construction of a surrogate explainable model capable of providing good estimations of the predictions of a black-box model \cite{charalampakos2024}. The surrogate model approximates the predictions of a reference black-box model (e.g., deep network models) with sufficient precision and/or fidelity, while being more explainable and interpretable. Although interesting and effective, this approach generally requires costly training processes and the availability of the training data of the reference black-box model. Changes in the structure of the input would require new (and costly) training episodes.
A key aspect of our approach is the ability to provide users with lightweight and flexible interpretations (e.g., changes to the semantic aggregation of the attributes) of the AI model without additional training processes. Furthermore, the use of a surrogate model is not strictly necessary in our case since we rely on a rule-based model that supports explainability by construction \cite{bologna2024}.

Other well-established techniques within the attribution-based methodology are SHAP \cite{shap} and LIME \cite{lime}. These techniques usually interpret model predictions in terms of an (explainable) linear function of binary variables whose sum approximates a prediction function of an AI model to explain/interpret. 
LIME \cite{lime} assumes that a single input might occur or not in a prediction. This binary interpretation does not account for ``layered'' or compound predictions where the same input could occur multiple times. The considered objective function estimating the original prediction is not directly applicable to our case since each prediction is made of multiple rules, and a single factor could occur multiple times. So, it is not easy to find a direct application of LIME without altering its ``standard'' use.
%
%

SHAP methods \cite{lipovetsky2001,strumbelj2014} require the possibility to retrain the model by removing specific attributes to evaluate their effects in the prediction. These methods aim to estimate SHAPley values \cite{Chen2023,shap} that approximate the effect of input attributes and support suitable interpretations.
Although well established, these methods were not feasible since we had limited access to the dataset, and couldn't consider retraining the model with all possible subsets of input attributes. 
%
Our objective is to design a method supporting a flexible interpretation of the outcome based on the level of expertise or profiles of target users. The combination of IR and semantic clustering of attributes enables a light, dynamic, and effective interpretation that can be easily adapted to different user needs. 
The taxonomical structure used to cluster attributes can be easily changed to specify different levels of details of interpretation and different granularity levels depending on the profile of the target user (e.g., AI developers would desire the highest level of details with a 1:1 mapping between attributes and factors). This level of flexibility would not be feasible with an interpretation mechanism requiring the model to be re-trained on an arbitrary subset of input attributes.

Other techniques require specific knowledge from domain experts to specify representative values of attributes for predicted classes to estimate their effects within a predictive model \cite{deeplift,bench2015}. Although promising, such approaches lack flexibility and strictly depend on static and accurate knowledge from experts, which is not always easy to capture and specify.
%
The work \cite{chen2022} proposes an additive risk model for lending decisions. They also propose attribute clustering and explanations based on the identification and ``weighting'' of risk factors. However, unlike our approach, the clustering is tightly coupled with the learning mechanism and shapes the structure of the underlying deep neural network. Our approach is also strictly correlated with the underlying prediction model \cite{bologna2024}, however, we keep interpretability distinct from the learning process. 
This separation allows us to adapt interpretations to the needs of target user profiles on the fly, without additional training iterations. That is, we can easily increase or decrease the level of detail of the interpretation by adapting the grain of the semantic clustering to the level of expertise of the target user.


Flexibility relies on a {\em semantic clustering} of the model's attributes to provide users with alternative views of the predictions. 
Each cluster determines the aggregation of the model's attributes that are semantically correlated with a certain {\em risk factor} and represents a factor that might somewhat affect the prediction. Interpretation correlates the attributes appearing in the conditions of the rules activated for a patient ({\em local rules}) with the associated factor. The analysis takes inspiration from IR metrics to compute the relevance of the factors.
In this regard, other approaches have investigated the combination of IR with explainability \cite{kyle2019,wan2023,paulsen2023}. 
%
%
%
%
However, the mentioned works use IR metrics to prepare the input of the AI model. IR metrics are thus mainly used to prepare the input vectors of the AI model by, for example, extracting words/tokens relevant for a prediction and/or classification task. 
To the best of the authors' knowledge, our approach originally uses IR metrics as post-hoc processing of predictions where tf-idf values actively contribute to the explainability and interpretability of predictions. The proposed approach provides target users with insights about the ``roles'' of input attributes without affecting the learning mechanisms in terms of accuracy or efficiency. 

%
%
%
%
\section{Conclusions and Future Works}
\label{sec:conclusions}
The explainability of AI-based predictive models is crucial to increase users' trust in the technology and foster their use in real-world decision-making processes. This is especially crucial in decision-critical domains like health. Furthermore, explainable AI models do not automatically imply the understandability of their outcomes from different types of users (stakeholders) \cite{ghassemi2021,combi2022}.
In this work, we have addressed the open challenge of enhancing the interpretability of an explainable AI-based model predicting the risk of arm lymphedema. Our approach adapts a standard statistical method from IR to the considered health domain to dynamically compute the relevance of risk factors determining the outcomes of the predictions.
A technical assessment demonstrates the feasibility of the approach and its ability to generate appropriate interpretations for both global and local explanations.
Additionally, a user-centered evaluation collects explicit feedback on the interpretability and usefulness of the approach from individuals with diverse areas of expertise (e.g., medicine, caregiving, AI development, data analysis).
Results show an increased interpretability of the explanation enriched with the score computation, compared to the raw explanation composed of the prediction rules only.
Future work will further investigate the explainability of the whole predictive framework developed and its concrete use within a clinical trial. In particular, we plan to ground the assessment on existing frameworks \cite{combi2022} that characterize the explainability from different perspectives (usability, usefulness, interpretability, understandability, etc.).

\bibliographystyle{unsrt}  
\bibliography{references}  

\begin{thebibliography}{10}

\bibitem{vellido2020}
Alfredo Vellido.
\newblock The importance of interpretability and visualization in machine learning for applications in medicine and health care.
\newblock {\em Neural Computing and Applications}, 32(24):18069--18083, 2020.

\bibitem{ghassemi2021}
Marzyeh Ghassemi, Luke Oakden-Rayner, and Andrew~L. Beam.
\newblock The false hope of current approaches to explainable artificial intelligence in health care.
\newblock {\em The Lancet Digital Health}, 3(11):e745--e750, 2021.

\bibitem{markus2021}
Aniek~F. Markus, Jan~A. Kors, and Peter~R. Rijnbeek.
\newblock The role of explainability in creating trustworthy artificial intelligence for health care: A comprehensive survey of the terminology, design choices, and evaluation strategies.
\newblock {\em Journal of Biomedical Informatics}, 113:103655, 2021.

\bibitem{Holzinger2019}
Andreas Holzinger, Georg Langs, Helmut Denk, Kurt Zatloukal, and Heimo Müller.
\newblock Causability and explainability of artificial intelligence in medicine.
\newblock {\em WIREs Data Mining and Knowledge Discovery}, 9(4), 2019.

\bibitem{Langer2021}
Markus Langer, Daniel Oster, Timo Speith, Holger Hermanns, Lena Kästner, Eva Schmidt, Andreas Sesing, and Kevin Baum.
\newblock What do we want from explainable artificial intelligence (xai)? – a stakeholder perspective on xai and a conceptual model guiding interdisciplinary xai research.
\newblock {\em Artificial Intelligence}, 296:103473, 2021.

\bibitem{Wang2019}
Fei Wang, Rainu Kaushal, and Dhruv Khullar.
\newblock Should health care demand interpretable artificial intelligence or accept "black box" medicine?
\newblock {\em Annals of internal medicine}, 172(1):59—60, January 2020.

\bibitem{tonekaboni2019}
Sana Tonekaboni, Shalmali Joshi, Melissa~D. McCradden, and Anna Goldenberg.
\newblock What clinicians want: Contextualizing explainable machine learning for clinical end use.
\newblock In Finale Doshi-Velez, Jim Fackler, Ken Jung, David Kale, Rajesh Ranganath, Byron Wallace, and Jenna Wiens, editors, {\em Proceedings of the 4th Machine Learning for Healthcare Conference}, volume 106 of {\em Proceedings of Machine Learning Research}, pages 359--380. PMLR, 09--10 Aug 2019.

\bibitem{xai2023}
Waddah Saeed and Christian Omlin.
\newblock Explainable ai (xai): A systematic meta-survey of current challenges and future opportunities.
\newblock {\em Knowledge-Based Systems}, 263:110273, 2023.

\bibitem{MONTAVON2018}
Grégoire Montavon, Wojciech Samek, and Klaus-Robert Müller.
\newblock Methods for interpreting and understanding deep neural networks.
\newblock {\em Digital Signal Processing}, 73:1--15, 2018.

\bibitem{combi2022}
Carlo Combi, Beatrice Amico, Riccardo Bellazzi, Andreas Holzinger, Jason~H. Moore, Marinka Zitnik, and John~H. Holmes.
\newblock A manifesto on explainability for artificial intelligence in medicine.
\newblock {\em Artificial Intelligence in Medicine}, 133:102423, 2022.

\bibitem{bologna2024}
Guido Bologna, Jean-Marc Boutay, Quentin Leblanc, and Damian Boquete.
\newblock Fidex: An algorithm for the explainability of ensembles and svms.
\newblock In Jos{\'e}~Manuel Ferr{\'a}ndez~Vicente, Mikel Val~Calvo, and Hojjat Adeli, editors, {\em Bioinspired Systems for Translational Applications: From Robotics to Social Engineering}, pages 378--388, Cham, 2024. Springer Nature Switzerland.

\bibitem{preact}
Foivos Charalampakos, Thomas Tsouparopoulos, Yiannis Papageorgiou, Guido Bologna, André Panisson, Alan Perotti, and Iordanis Koutsopoulos.
\newblock Research challenges in trustworthy artificial intelligence and computing for health: The case of the pre-act project.
\newblock In {\em 2023 Joint European Conference on Networks and Communications \& 6G Summit (EuCNC/6G Summit)}, pages 629--634, 2023.

\bibitem{amann2020}
Julia Amann, Alessandro Blasimme, Effy Vayena, Dietmar Frey, Vince~I. Madai, and the~Precise4Q consortium.
\newblock Explainability for artificial intelligence in healthcare: a multidisciplinary perspective.
\newblock {\em BMC Medical Informatics and Decision Making}, 20(1):310, 2020.

\bibitem{RibeiroPereira2017}
Ana Carolina~Padula Ribeiro~Pereira, Rosalina~Jorge Koifman, and Anke Bergmann.
\newblock Incidence and risk factors of lymphedema after breast cancer treatment: 10 years of follow-up.
\newblock {\em The Breast}, 36:67--73, 2017.

\bibitem{park2017}
Jin~Hee Park, Won~Hee Lee, and Hae~Soo Chung.
\newblock Incidence and risk factors of breast cancer lymphoedema.
\newblock {\em Journal of Clinical Nursing}, 17(11):1450--1459, 2008.

\bibitem{Faul2009}
Franz Faul, Edgar Erdfelder, Axel Buchner, and Albert-Georg Lang.
\newblock Statistical power analyses using g*power 3.1: Tests for correlation and regression analyses.
\newblock {\em Behavior Research Methods}, 41:1149--1160, 2009.

\bibitem{gdpr2016}
{European Union}.
\newblock {Regulation (EU) 2016/679 of the European Parliament and of the Council of 27 April 2016 on the protection of natural persons with regard to the processing of personal data and on the free movement of such data (General Data Protection Regulation)}.
\newblock \url{https://eur-lex.europa.eu/eli/reg/2016/679/oj}, 2016.
\newblock Official Journal of the European Union, L119, 1--88.

\bibitem{charalampakos2024}
Foivos Charalampakos and Iordanis Koutsopoulos.
\newblock Exploring multi-task learning for explainability.
\newblock In S{\l}awomir Nowaczyk, Przemys{\l}aw Biecek, Neo~Christopher Chung, Mauro Vallati, Pawe{\l} Skruch, Joanna Jaworek-Korjakowska, Simon Parkinson, Alexandros Nikitas, Martin Atzm{\"u}ller, Tom{\'a}{\v{s}} Kliegr, Ute Schmid, Szymon Bobek, Nada Lavrac, Marieke Peeters, Roland van Dierendonck, Saskia Robben, Eunika Mercier-Laurent, G{\"u}lg{\"u}n Kayakutlu, Mieczyslaw~Lech Owoc, Karl Mason, Abdul Wahid, Pierangela Bruno, Francesco Calimeri, Francesco Cauteruccio, Giorgio Terracina, Diedrich Wolter, Jochen~L. Leidner, Michael Kohlhase, and Vania Dimitrova, editors, {\em Artificial Intelligence. ECAI 2023 International Workshops}, pages 349--365, Cham, 2024. Springer Nature Switzerland.

\bibitem{shap}
Scott~M. Lundberg and Su-In Lee.
\newblock A unified approach to interpreting model predictions.
\newblock In {\em Proceedings of the 31st International Conference on Neural Information Processing Systems}, NIPS'17, page 4768–4777, Red Hook, NY, USA, 2017. Curran Associates Inc.

\bibitem{lime}
Marco~Tulio Ribeiro, Sameer Singh, and Carlos Guestrin.
\newblock {``Why Should I Trust You?'': Explaining the Predictions of Any Classifier}.
\newblock In {\em Proceedings of the 22nd ACM SIGKDD International Conference on Knowledge Discovery and Data Mining}, KDD '16, page 1135–1144, New York, NY, USA, 2016. Association for Computing Machinery.

\bibitem{lipovetsky2001}
Stan Lipovetsky and Michael Conklin.
\newblock Analysis of regression in game theory approach.
\newblock {\em Applied Stochastic Models in Business and Industry}, 17(4):319--330, 2001.

\bibitem{strumbelj2014}
Erik {\v{S}}trumbelj and Igor Kononenko.
\newblock Explaining prediction models and individual predictions with feature contributions.
\newblock {\em Knowledge and Information Systems}, 41(3):647--665, Dec 2014.

\bibitem{Chen2023}
Hugh Chen, Ian~C. Covert, Scott~M. Lundberg, and Su-In Lee.
\newblock Algorithms to estimate shapley value feature attributions.
\newblock {\em Nature Machine Intelligence}, 5(6):590--601, Jun 2023.

\bibitem{deeplift}
Avanti Shrikumar, Peyton Greenside, and Anshul Kundaje.
\newblock Learning important features through propagating activation differences.
\newblock In {\em Proceedings of the 34th International Conference on Machine Learning - Volume 70}, ICML'17, page 3145–3153. JMLR.org, 2017.

\bibitem{bench2015}
Sebastian Bach, Alexander Binder, Grégoire Montavon, Frederick Klauschen, Klaus-Robert Müller, and Wojciech Samek.
\newblock On pixel-wise explanations for non-linear classifier decisions by layer-wise relevance propagation.
\newblock {\em PLOS ONE}, 10(7):1--46, 07 2015.

\bibitem{chen2022}
Chaofan Chen, Kangcheng Lin, Cynthia Rudin, Yaron Shaposhnik, Sijia Wang, and Tong Wang.
\newblock A holistic approach to interpretability in financial lending: Models, visualizations, and summary-explanations.
\newblock {\em Decision Support Systems}, 152:113647, 2022.

\bibitem{kyle2019}
Kyle Martin, Anne Liret, Nirmalie Wiratunga, Gilbert Owusu, and Mathias Kern.
\newblock Developing a catalogue of explainability methods to support expert and non-expert users.
\newblock In Max Bramer and Miltos Petridis, editors, {\em Artificial Intelligence XXXVI}, pages 309--324, Cham, 2019. Springer International Publishing.

\bibitem{wan2023}
Wan~Yit Yong, Rajesh Jaiswal, and Fernando Perez~Tellez.
\newblock Explainability in nlp model: Detection of covid-19 twitter fake news.
\newblock In {\em Proceedings of the 2023 Conference on Human Centered Artificial Intelligence: Education and Practice}, HCAIep '23, page~1, New York, NY, USA, 2023. Association for Computing Machinery.

\bibitem{paulsen2023}
Derek Paulsen, Yash Govind, and AnHai Doan.
\newblock Sparkly: A simple yet surprisingly strong tf/idf blocker for entity matching.
\newblock {\em Proc. VLDB Endow.}, 16(6):1507–1519, 2023.

\end{thebibliography}

\end{document}